\documentclass{article}

\usepackage{arxiv}

\usepackage[utf8]{inputenc} 
\usepackage[T1]{fontenc}    
\usepackage{hyperref}       
\usepackage{url}            
\usepackage{booktabs}       
\usepackage{amsfonts}       
\usepackage{nicefrac}       
\usepackage{microtype}      
\usepackage{lipsum}
\usepackage{graphicx}
\graphicspath{ {./images/} }

\title{ATP: A holistic attention integrated approach to enhance ABSA}

\author{
 Ashish Kumar \\
  School of Computer and Systems Sciences\\
  Jawaharlal Nehru University\\
  New Delhi-110067, India \\
  \texttt{ashishkumar2912@gmail.com} \\
   \And
 Vasundhra Dahiya \\
  Digital Humanities\\
  Indian Institute of Technology\\
  Jodhpur-342037, India \\
  \texttt{vasundhra.dahiya@gmail.com} \\
  \And
 Aditi Sharan \\
  School of Computer and Systems Sciences\\
  Jawaharlal Nehru University\\
  New Delhi-110067, India \\
  \texttt{aditisharan@gmail.com} \\
}

\begin{document}
\maketitle
\begin{abstract}
Aspect based sentiment analysis (ABSA) deals with the identification of the sentiment polarity of a review sentence towards a given aspect. Deep Learning sequential models like RNN, LSTM, and GRU are current state-of-the-art methods for inferring the sentiment polarity. These methods work well to capture the contextual relationship between the words of a review sentence. However, these methods are insignificant in capturing long-term dependencies. Attention mechanism plays a significant role by focusing only on the most crucial part of the sentence. 
In the case of ABSA, aspect position plays a vital role. Words near to aspect contribute more while determining the sentiment towards the aspect. Therefore, we propose a method that captures the position based information using dependency parsing tree and helps attention mechanism. Using this type of position information over a simple word-distance-based position enhances the deep learning model's performance. We performed the experiments on SemEval'14 dataset to demonstrate the effect of dependency parsing relation-based attention for ABSA.
\end{abstract}


\section{Introduction\label{ATP_intro}}
Sentiment Analysis is the process of mining people's opinions regarding a specific product, service, organization, etc. The task of sentiment analysis provides various benefits to producers as well as to consumers also. It helps producers improve the quality of their products and helps consumers buy the best product by analysis of reviews.
Sentiment analysis has various granularities depending on the text they are focusing on. If a whole document (e.g. review) is considered while examining the sentiment associated with it, it is called document-level sentiment analysis. Similarly, if the sentiment of each sentence of review is extracted, it is called sentence-level sentiment analysis. 
There is an assumption that focused text contains only uni-sentiment.
But at the practical level, a single sentence can talk about an entity's different aspects (features), and users can have different sentiment orientations towards various aspects.
Therefore, tackling the problem of sentiment analysis should have different perceptive that should take care of the aforesaid practical constraints.

Aspect-based sentiment analysis (ABSA) is a fine-grained analysis of sentiment present in the text\cite{liu2015sentiment}.
ABSA works by dealing with different aspects of an entity and determines whether the sentiment is positive or negative concerning each aspect. In order to understand this notion, let us consider the following example:
\textit{``Looks nice, but has a horribly cheap feel.''}
This excerpt from a customer review contains \textit{positive} sentiment towards aspect-term \emph{looks} but on the other hand, \textit{negative} sentiment towards aspect-term \emph{feel}.
Tackling the sentiment analysis problem should have a divergent perspective that should take care of the aforesaid practical constraints. The sentiment polarity of a sentence is exceptionally subject to both context and target. 
One of the tasks of ABSA is to extract the sentiment polarity of a target i.e. aspect-term in any given sentence. 
In this article, we use target and aspect-term interchangeably.

Different context words have various effects on deciding the notion of sentiment orientation of text towards the aspect-target. To recognize the sentiment of an individual aspect-target, one basic undertaking is to mold relevant contextual features for the aspect-target in the original text. 
In the process of determining the sentiment polarity of a review sentence towards a target, target information should be taken into account efficiently. 
There is some research\cite{tang2015target} those advocates the consideration of aspect-target information in the determination of target-dependent sentiment.
Apart from target information, as claimed by Gu et al. (2018)\cite{gu2018position} neighboring words also have a more significant contribution to investigate the sentiment orientation of an aspect-term. For examples:

Review 1:
\textit{``Keyboard is great, very quiet for all the typing that I do.''}

Review 2:
\textit{``Fresh ingredients and everything is made to order.''}
\\

In Review 1, word `great'  neighboring to aspect-term \emph{keyboard} determines its sentiment polarity while in Review 2, words `fresh' neighboring to aspect-term \emph{ingredients} determines the sentiment polarities.
However, in some cases, words at a long distance based on word-distance have significant contributions also. 

Considering the state of the art deep learning approaches used for ABSA, we found some research gaps that most of the techniques only use neighboring words to handle the context while others focus on complete sentences. As a result, an irrelevant portion of text dominates in prediction time. 
Focusing on these research gaps, we proposed Bi-LSTM attention based model to perform the Aspect-term sentiment analysis (ATSA). In the case of ATSA, the position of the opinion-word with respect to the aspect-term plays a vital role. Opinion-words near to aspect-term contribute more while determining the sentiment towards that aspect-term. In the earlier works, the distance has been captured by the number of words between the aspect-terms and opinion-words. However, this criteria may not be sufficient to capture the dependence of opinion and the aspect-terms. The syntax of the sentence should also be taken into consideration while determining the distance between aspect-terms and opinion-words. This motivates capturing the position of the opinion-word with respect to the aspect-term using a dependency parsing tree and helps the attention mechanism. Using this type of position information over a simple word-distance based position enhances the performance of deep learning models. We are motivated to handle these issues and decided to use the aforementioned information in a systematization way to overcome existing models. In this line of research, we have composed a model named ATP that first generates sentence representation using Bi-LSTM. After that attention mechanism is applied to these hidden representations. To propagate the aspect-term specific sentiment context, dependency tree would be propitious. Hence, the speciality of this attention mechanism is that it focuses on the important context using aspect-term and aspect-term related dependency-tree path information.
The key contributions of our work are summarized as follows:
\begin{enumerate}

 \item We proposed a deep learning architecture (Bi-LSTM) to identify the sentiment polarity of a review sentence, with respect to, an aspect-target.
 \item We have used the attention mechanism to focus only on the relevant part of the sentence instead of focusing on a complete sentence while calculating sentiment.
 \item We have utilized the dependency parsing tree-based position information to get more attention on context that is strongly allied with aspect-target.
 \item We have also focused on the sentences having the conflict sentiment polarity, apart from positive, negative, and neutral sentiments.
 \item Empirical results on a benchmark dataset show the significance of the role of dependency parsing relation based information in ABSA.
\end{enumerate}

The remainder of the article is organized as follows:
Section~\ref{ATP_rw} highlight some research on sentiment analysis. 
In Section~\ref{ATP_bm}, background and motivation are presented followed by
Section~\ref{ATP_dep} includes dependency parsing relation.
In Section~\ref{ATP_mod_desc}, model description is presented. 
Section~\ref{ATP_exp} deals with experimental settings and results. Finally, the article is concluded in Section~\ref{ATP_con}.

\section{Related Work\label{ATP_rw}}
Initially, sentiment classification task was done on document or sentence level\cite{pang2002thumbs}. Sentiment orientation of text was calculated by finding sentiment bearing words (opinion words) from the text and assigning sentiment score according to their orientation (i.e. positive or negative). Further sentiment scores of all opinion words were aggregated to find out the overall sentiment orientation.
To detect opinion words different sentiment lexicons i.e. WordNet\cite{hu2004mining}, SenticNet\cite{cambria2010senticnet}, SentiFul\cite{neviarouskaya2011sentiful} etc. were used. These sentiment lexicons are dictionary type structure that contains a variety of opinion words and phrases along with their sentiment scores. It has been observed that generally, opinion words are adjective or adverb, so these words were also treated as candidate opinion words in some work\cite{qiu2009expanding}.

Aspect-based sentiment classification problem is also treated as a supervised classification problem. Traditional classification algorithms like SVM\cite{yu2011aspect}, CRF\cite{li2010structure}, HMM\cite{jin2009opinionminer} were applied to solve this problem. These algorithms work on manually drafted features, so the performance of these algorithms depends on how efficiently these features are designed.

With the advent of deep learning methods, text representation was done as distributed representations of words in a vector space\cite{mikolov2013distributed}. This representation encodes many linguistic regularities and patterns, helping in achieving good performance in various NLP tasks. These representations are called word-embeddings.
The invention of word-embeddings motivates researchers to apply deep learning methods in the field of sentiment analysis\cite{zhang2018deep}. 
Researchers apply various sequential models like RNN, LSTM, GRU to infer the sentiment polarity. These models are capable enough to capture the intricate relationship between the context words of text without any feature engineering.

Aspect based information is also passed in the deep-learning models to detect the sentiment with respect to given aspect\cite{tang2016aspect}. This aspect level information was passed in the form of aspect-embeddings which was calculated as an average of word-embeddings of all the words composing the aspect.

While calculating sentiment polarity, all words are not equally important; there are some words that have more influence. So in practice, these words should be given more importance while calculating sentiment polarity. 
There is a mechanism called the attention mechanism, which knows ``where to look''. It assigns a weight to each context word according to their importance. After the success of attention mechanism in machine translation task\cite{bahdanau2014neural}, it is applied in the field of ABSA and provided promising results\cite{wang2016attention}.
Gu et al. (2018)\cite{gu2018position} proposed a position-aware bidirectional attention network (PBAN) based on bidirectional Gated Recurrent Units (Bi-GRU) for aspect level sentiment classification. Along with aspect-embeddings, this model uses position-embeddings. Position embeddings were generated with the use of word distance with the aspect-term.

Our work is slightly different from Gu et al. (2018)\cite{gu2018position} in terms of calculation of position-embeddings. We are using dependency tree information to calculate position embeddings.
We believe that a word adjacent to aspect-term in form of dependency path length in-spite of having long word-distance has more influence. 

\section{Background and Motivation\label{ATP_bm}}
One of the applications which use a lexicon-based approach is Sentiment analysis. Here, sentiment lexicons are created in order to provide sentiment scores to the text. A dictionary-like structure defines this lexicon; by using words and phrases of the given text as sentiment words and has sentiment polarities or sentiment scores for those words. A challenge is that this approach follows a plurality of context and domain, i.e., different sentiment orientations in different domains. For e.g.: 'funny movie' is the opposite of 'funny taste' where the sentiment word is the same but not the sentiment polarity—the polarity changes as the domain changes from movie to food\cite{kumar2021ate}. The failure to capture context and domain sensitivities is one limitation of lexicons. Deep learning captures context-sensitivity well and avoids the extensive step of labeling sentiment words.

Only sentence specific information is not enough to capture the sentiment polarity of a sentence regarding aspect-terms presented in the sentence. Many researchers tried to use aspect-terms based information along with sentence information while capturing sentiment polarity. Using aspect-terms information strengthens the performance of the task. Some researchers split the sentence according to the aspect-terms\cite{tang2015effective} while others use aspect-terms information in form of embedding concatenation with sentence embeddings or with hidden representations of sentence\cite{wang2016attention}. In spite of increasing the performance of models, these models lack capturing the important context in the sentence given some aspect-terms. Thus, it fails to capture the sentiment of sentences such as: \textit{``Waiting three hours before getting our entrees was a treat as well.''} To overcome this problem research community focus to attend the prominence opinion determining context regarding given aspect-terms using attention mechanism. 

Opinion-words that are syntactically nearby aspect-target contribute more to the detection of sentiment orientation. But this nearby is relative, not absolute. Therefore, we try to capture the relative word position (dependency tree path-distance) instead of absolute word position (word-distance). Since review text follows the grammar rule, dependency parsing would be beneficial (But for irregular grammar text like tweets, etc. dependency parsing would not be an option). Considering a review:

\textit{``We had pizza and despite making a mess, it was a ton of fun and quite tasty as well.''}

\begin{figure}
 \centering
 \includegraphics[width=0.8\linewidth]{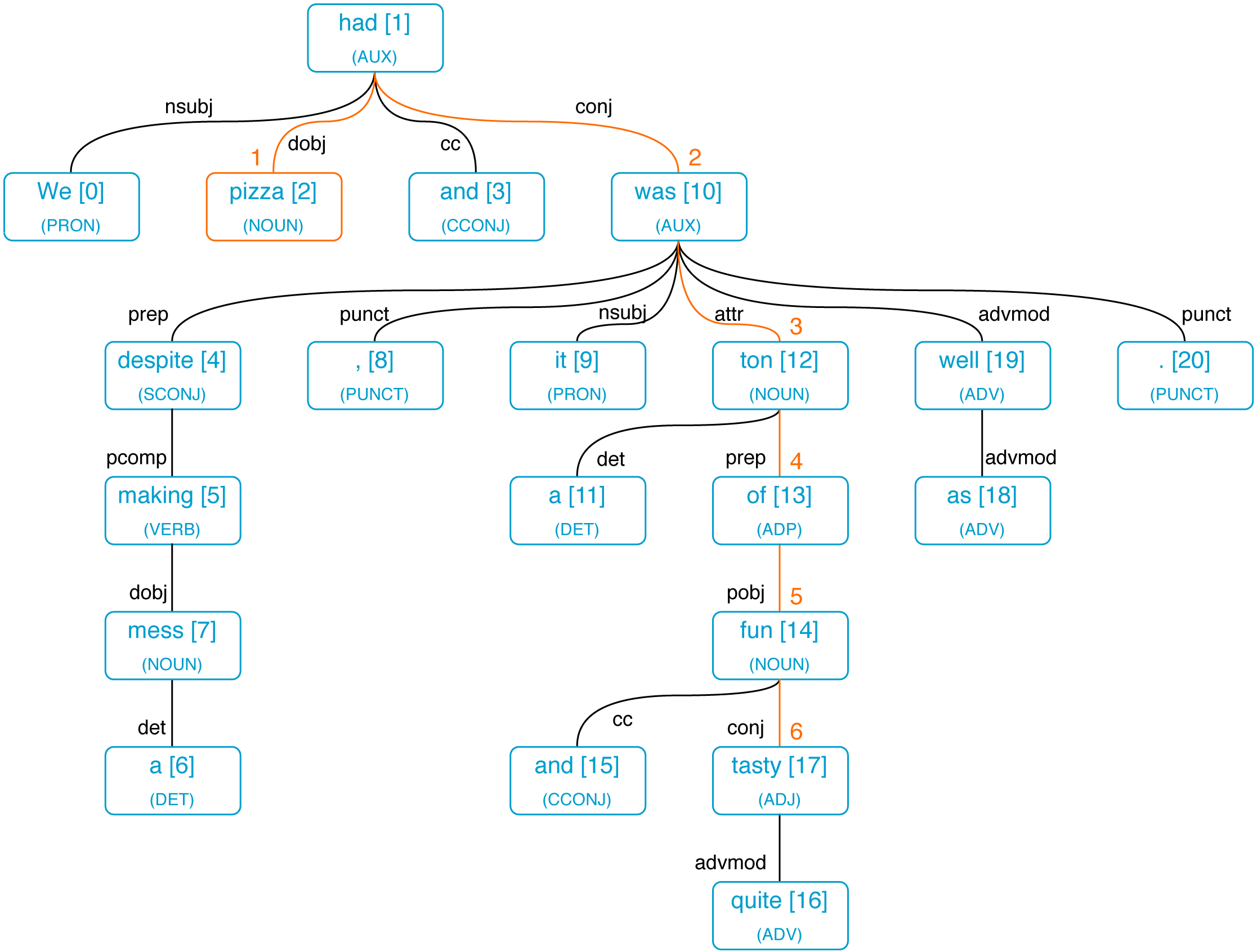}
 \caption{Example of dependency path distance}
 \label{FIG:ATP_dep_graph_dist}
\end{figure}

As in the mentioned review, a word \textit{tasty} is far based on the word-distance but near based on the dependency path-distance.
In this example, word \textit{tasty} is the main determiner of sentiment orientation of aspect-term \emph{pizza}.
This concludes that words having short dependency path-distance to target word irrespective to their word-distance, are more important to target word while determining the sentiment polarity towards that target.
Figure~\ref{FIG:ATP_dep_graph_dist} shows the dependency path-distance of \textit{tasty} with respect to aspect-term \emph{pizza}.

In sequential deep learning approaches, attention mechanism has the ability to learn the text representation by paying more attention to a significant portion of the text. There are some attention-based models used in sentiment analysis tasks too \cite{he2018effective,li2018aspect}. Such mechanism takes an external memory and representation of a sequence as input and produces a probability distribution quantifying the concerns in each position of the sequence. In the work of Gu et al. (2018)\cite{gu2018position}, they proposed a position-aware bi-directional network (PBAN) for ABSA. This model utilized the position based embeddings for calculating attention weights. We borrowed the idea of position-based embeddings from the aforesaid architecture. Since dependency parsing trees are good for capturing the words relationship in a sentence, we have also incorporated dependency information to calculate the position-based embedding.

\section{Dependency Parsing Relation\label{ATP_dep}}
\begin{figure}[!ht]
 \centering
 \includegraphics[width=0.7\linewidth, height=2.5cm]{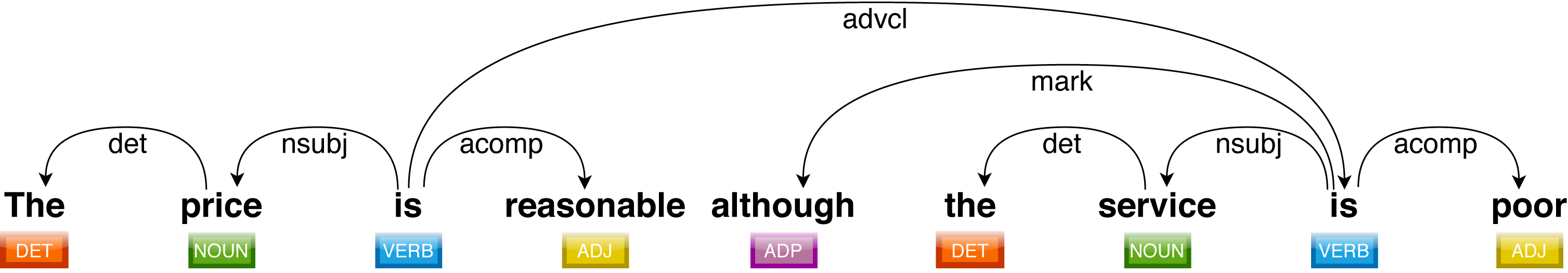}
 \caption{Example of the dependency parsing graph}
 \label{FIG:dependencyGraph}
\end{figure}

One of the most widely used techniques to assign syntactic structures to sentences are Parse trees.
This is formally known as Syntactic parsing or Dependency parsing. These are formed using some parsing algorithms.
This involves extracting a dependency parse of a sentence in such a way that the sentence's grammatical structure is captured.
This structure describes the grammatical relationship between words of the sentence in form of relation triplets.
This relation between pairs of words, these dependencies are mapped onto a directed graph where words become nodes and grammatical relations become edge labels of the graph. 
Dependency parsing proves beneficial for tasks like Semantic Analysis and Grammar Checking. Figure~\ref{FIG:dependencyGraph} demonstrates the dependency graph on a review sentence.

\section{Model Description\label{ATP_mod_desc}}
This section covers the model description followed by subsections detailing problem statement, input representation, and the model's architecture.

\subsection{Problem Statement}
Given a review sentence $R = \{w_1, w_2, \dots,w_n\}$ consisting $n$ words, and aspect-target $T= {t_1, t_2, \dots, t_m}$ that is a continued $m$ words sub-sequence in the review sentence. The task is to determine the sentiment polarity $p_i$ of aspect-target $T$ towards the review sentence $R$ from a polarity class set $P$.

\subsection{Input Representation}
\subsubsection{Text Representation (Word-embeddings)}
Traditional word-embeddings like Word2Vec and GloVe were the pioneer text representations to capture the semantic of the text. These representations are very helpful to determine various word-analogies since they were trained on a large dataset. But generally, fails to differentiate the meaning of polysemous words because in Word2Vec and GloVe a unique word always has a unique representation irrespective of their meaning or context they have used in. So, there is a need for context-dependent word representations. These are called pre-trained embeddings. ELMo, BERT are some examples of these embeddings. We have used ELMo in our model. ELMo (Embeddings from Language Model) provide contextualized word representation. It takes the full sentence as an input and generates the word-representation (embedding). The learning task of ELMo is to predict the next word through a deep learning model (BiLSTM) given a sequence of words. Since they also use character-based representation, thus are really helpful to generate the embeddings in case of out-of-vocabulary words.

\subsubsection{Aspect-term Representation}
Using the concept of word-embeddings any word can be mapped into a dense vector representation. An aspect-term can be a single word or a phrase that consists of many words. For the cases, when an aspect-term is a single word, it can be represented by the embeddings of that word. However, if the aspect-term is a phrase like screen resolution then it can be represented by averaging all the constituting words' word-embeddings\cite{tang2016aspect}. In our model, we have used the aspect-terms in the same way as input without averaging the aspect-terms.
 
\subsubsection{Position Representation}
Position information of target aspect-term is modeled using the dependency parse tree. The dependency path length of the word from a target is used as a position distance. In this way, words directly connected with target aspect-term share the same position distance that is 1. This position distance value increases with the increase in path length. For the given example ``\textit{The price is reasonable although the service is poor}'' position distance vectors will be $P=[1, 0, 1, 2, 3, 4, 3, 2, 3]$ and $P=[4, 3, 2, 3, 2, 1, 0, 1, 2]$ for aspect-term \emph{price} and \emph{service} respectively (refer Figure~\ref{FIG:dependencyGraph}). 
These position distance values of position vector will be replaced by position embeddings which will be obtained by a position embedding matrix $P\in \mathbb{R}^{d_p \times N}$, 
$d_p$ indicates the embedding dimension and $N$ denotes maximum sentence length. 

\subsection{Model Architecture}
In this section, we discuss the deep learning based sequential model that we have used in our proposed method. First, we describe the basic Bi-LSTM followed by Bi-LSTM with attention. The motivation of adding attention into Bi-LSTM is that standard Bi-LSTM processes the input in a sequential way and performs the same operation on each input word. Due to this nature, it is unable to detect the importance of each context word. According to Tang et al. (2016b)\cite{tang2016aspect}, only some part of the context has more influence on aspect-target and assists to infer the sentiment orientation. But which subset of context should model focus on is unknown and varies sentence by sentence. A mechanism that solves this problem is attention mechanism. It is capable of finding out the importance of each word in a sentence given a target word. Following, we explain how Bi-LSTM and Attention mechanisms work inferring how it helps our case.

\subsubsection{Bi-LSTM}
Traditionally, RNNs are invented to handle sequential data. But these networks suffer from the exploding and vanishing gradient problem. Due to this, networks' weight changes rapidly (exploding gradient) or not updated/dropped (vanishing gradient). To handle the problem of vanishing gradients different variations of RNN were proposed, one of them being LSTM. A standard LSTM uses the concept of gates to maintain the memory states. These gates control the flow of information in memory by helping when to add new information (input gate $i$), when to update or forget old information (forget gate $f$), and what to output conditioned on input and memory cell content (output gate $o$).

In Bi-LSTM, two LSTMs are used. The cell update process of both LSTM (forward and backward) is the same, with one difference being direction. Forward LSTM receives the input in the same sequence they appear while backward LSTM receives the input in the reversed sequence. The final output of Bi-LSTM cell is the concatenation of hidden activation states produced by both LSTMs, as shown in Eq.~\ref{ATP_eq_1}.
\begin{equation}
\label{ATP_eq_1}
h^{<t>} = [\overrightarrow{h}^{<t>};\overleftarrow{h}^{<t>}] \in \mathbb{R}^{d_h+d_h}
\end{equation}
Where $\overrightarrow{h}^{<t>}$, $\overleftarrow{h}^{<t>}$ are the output states of each LSTM at time-step $t$.

\subsubsection{Bi-LSTM with Attention}
The encoding process of text deems senseless if a sequence too long is input. The encoded vector must capture complete information, consequently, the meaning of the sequence, but this might fail if the sequence is too long. Bahdanau et al. (2014)\cite{bahdanau2014neural} offered the 'Attention' mechanism to resolve this. Through attention, the focus is assured on some sentence parts. In tasks such as translation, context capturing is relevant, and attention captures it successfully, even if a sentence stretches throughout a long sequence. In the proposed framework, the context vector $c$  deploys the attention mechanism. 
It is defined by a weighted sum of features at different time-steps $h^{<t>}$ across the input given by Eq.~\ref{ATP_eq_2}.
\begin{equation}
\label{ATP_eq_2}
c = \sum_{t=1}^{T_x} \alpha^{<t>} h^{<t>}
\end{equation}

Here $\alpha$ is the attention parameter that depicts the amount of attention to be applied on different features. Total sum of all the attention weight should be equal to 1 i.e. $\sum_{t=1}^{T_x} \alpha^{<t>} = 1$

The attention weight $\alpha^{<t>}$ is determined by Eq.~\ref{ATP_eq_4}, \ref{ATP_eq_5}.

\begin{equation}
\label{ATP_eq_4}
\alpha^{<t>} = \frac{\exp(score(h^{<t>},p^{<t>},a^{<t>}))}{\sum_{t=1}^{T_x} \exp(score(h^{<t>},p^{<t>},a^{<t>}))} 
\end{equation}

\begin{equation}
\label{ATP_eq_5}
score(h^{<t>},p^{<t>},a^{<t>}) = W_s\tanh(W_{hpa} [h^{<t>};p^{<t>};a^{<t>}]
\end{equation}
where, 
$W_s \in \mathbb{R}^{1 \times d_w}$,
$W_{hpa} \in \mathbb{R}^{d_w \times (d_e+2d_h+d_p)}$, and inputs to score function are Bi-LSTM output $h^{<t>}\in \mathbb{R}^{2d_h}$, aspect-position vector $p^{<t>}\in \mathbb{R}^{d_p}$, aspect-terms embedding $a^{<t>}\in \mathbb{R}^{d_e}$

\begin{figure}[!ht]
\centering
\includegraphics[width=0.8\linewidth]{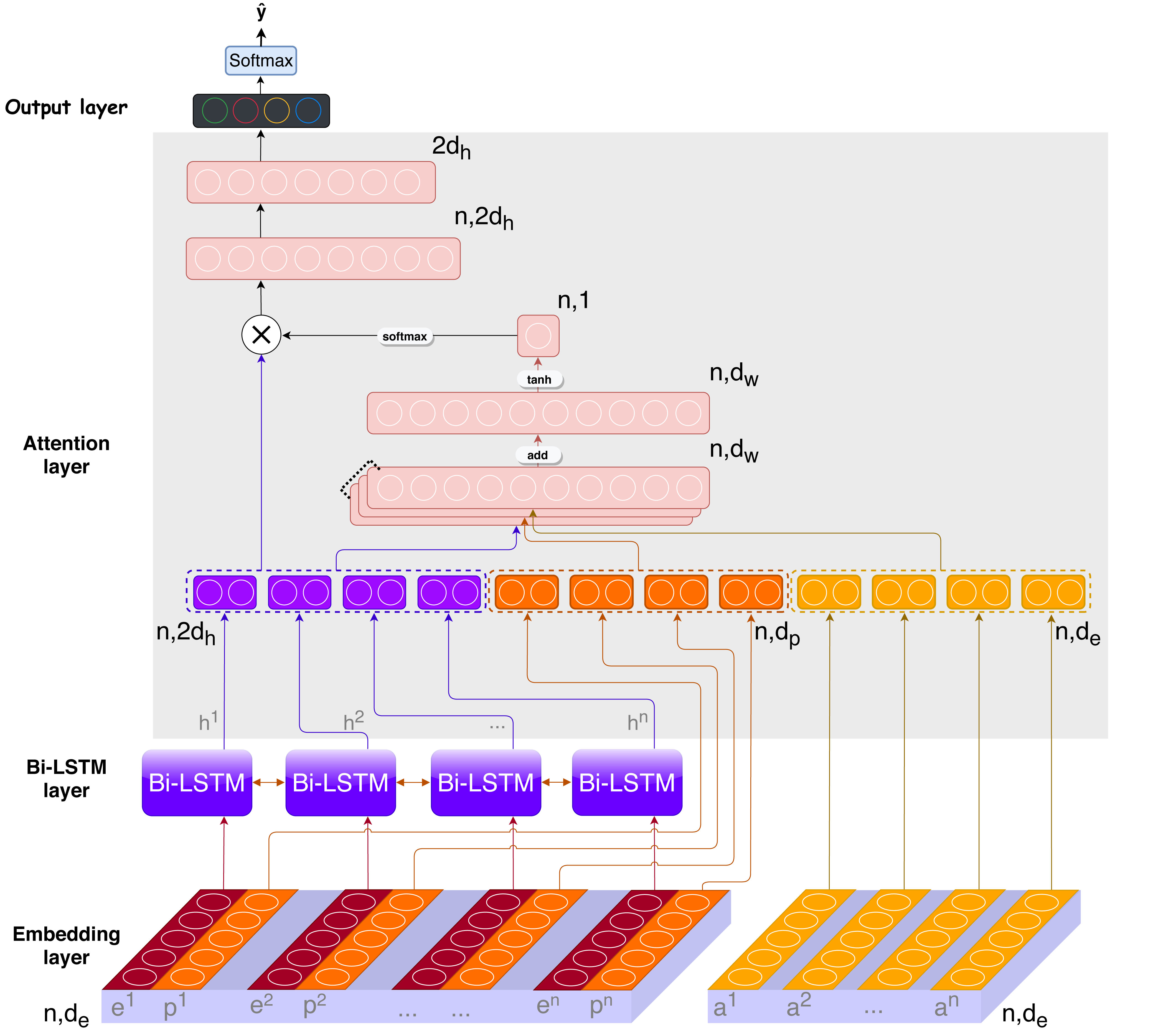}
\caption{ATP model architecture}
\label{FIG:ATP_model}
\end{figure}

The whole process of attention is integrated with the model architecture that jointly learns the attention parameters along with the model's parameters. Notably, in spite of producing a common fixed-length vector for a particular input sequence, the model generates aspect-dependent context vectors that focus on various sections of the input sequence given for the focused aspect-term. Figure~\ref{FIG:ATP_model} shows the architecture of our proposed model.

\section{Experimental Settings and Results\label{ATP_exp}}
In this section, we discuss the dataset, base models, parameter settings, model training and analyze the experimental results performed for ATSA.

\subsection{Base Models}
In this section, we have discussed base models that are used to compare our method. 

\begin{enumerate}
 \item \textbf{TD-LSTM:} Tang et al. (2016a)\cite{tang2015effective} uses target-dependency as a criterion for analyzing sentences. Under the assumption that words related to the target word comprise context required for identifying polarity sentences. To scan for context, they used BiLSTMs. Context captured from both directions (preceding context plus target word and target word plus following context) is concatenated. This target-specific text vector representation is passed to the softmax layer which outputs the sentiment polarity. Even with capturing target-dependent context from the sentence, the limitation that could be noticed with this approach is the lack of covering semantic relationships between context and aspect target word.

 \item \textbf{AE-LSTM, ATAE-LSTM:} Wang et al. (2016)\cite{wang2016attention} follows an interesting attention-based approach. It implements simple LSTMs for modeling context-words and then combines these embeddings with target aspect-embeddings to generate attention-vectors. This initially used approach is only an Attention-based LSTM (AT-LSTM). Later, an Attention-based LSTM with Aspect Embedding (ATAE-LSTM) is used. The difference is that the former concatenates aspect embeddings with the output of LSTM layer to be passed into the attention layer. The latter, however, the aspect embeddings are combined at each input of LSTM. Thus, not only accommodates the aspect embeddings for better sentiment identification but the mutual relationship of these embeddings with the context word vector. This provides much better contextualized embeddings.

 \item \textbf{MemNet:} For ABSA, Tang et al. (2016b)\cite{tang2016aspect} uses Memory Networks. They are popularly used with Question Answering tasks. For ABSA, capturing the importance of context words using Memory neural networks (MemNet) is achieved by treating context-words as factual description and aspect-terms as the query itself. Here, an external memory stacks context word-vectors. As MemNet applies attention multiple times, the last layer is already ready for prediction without any secondary concatenation, it fundamentally achieves aspect-level sentiment classification. In this memory network, an external memory space is made by stacking the context word vectors. With a computational layer named `hop', an aspect vector was taken as input, and then, using the attention mechanism, memory was focused on. Every hop returned a sum of linearly transformed aspect vector and attention layer output. This sum is passed to the next layer. The output of the last hop is considered as sentence representation, with respect to an aspect which is at last passed through softmax classifier for sentiment classification.

 \item \textbf{IAN:} Ma et al. (2017)\cite{ma2017interactive} uses two components (one for target and another for context representation) composed from LSTM that interact with each other to provide attention to other components. Such interactive-attentive mechanism covers both contextual and semantic relationships between the aspect-term and its context.

 \item \textbf{RAM:} Chen et al. (2017a)\cite{chen2017recurrent} introduced a model RAM that essentially used multiple attention mechanisms to produce informative aspect-dependent sentiment representation. It generates memory using Bi-LSTM. This location weighted memory comprises the relative position of words with respect to aspect target. A recurrent network (GRU) was used to capture multiple attention on memory. Finally, the softmax classifier was used for sentiment classification.

 \item \textbf{PBAN:} Gu et al. (2018)\cite{gu2018position} proposed a position-aware bidirectional attention network (PBAN) based on bidirectional Gated Recurrent Units (Bi-GRU) for aspect level sentiment classification. Along with aspect-embeddings, this model uses position-embeddings. Position embeddings were generated with the use of word distance with the aspect-term.

 \item \textbf{Coattention-MemNet:} Yang et al. (2019)\cite{yang2019aspect} proposed an alternating co-attention mechanism to process both target and context to reduce the noise from target-representation. Further, a Coattention-Memory network was utilized to infer the sentiment polarity.

 \item \textbf{GANN:} Liu, Shen (2020a)\cite{liu2020aspect} proposed a combination of RNN and CNN, a language-independent model called Gated Alternate Neural Network (GANN) that uses various sentiment clues like relative distance between aspect target and context word, sequence information, and semantic dependency.

 \item \textbf{ReMemNN:} Liu, Shen (2020b)\cite{liu2020rememnn} used pre-trained embeddings, BERT. The neural model to process was ReMemNN i.e. Recurrent Memory Neural Network. It is basically an attempt to overcome the problem of pre-trained word-embeddings using the embedding adjustment learning module in the field of sentiment analysis. They have also incorporated a multielement attention mechanism for a more precise aspect dependent sentiment analysis.

\end{enumerate}
\subsection{Dataset}
\begin{figure}
 \centering
 \includegraphics[width=0.8\linewidth]{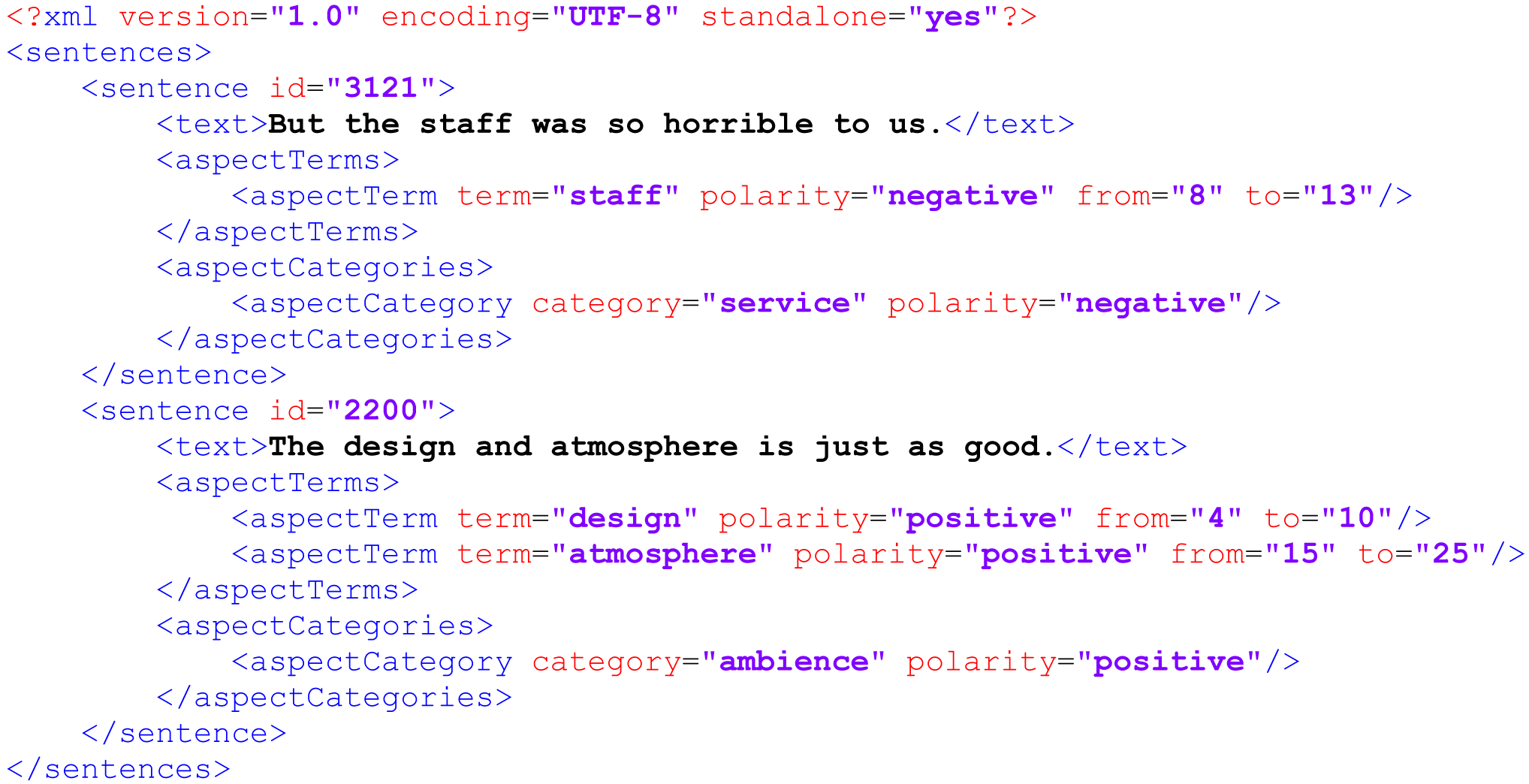}
 \caption{An XML snippet of dataset}
 \label{FIG:SemEval14Dataset}
\end{figure}

The International Workshop on Semantic Evaluation series (SemEval-2014, 2015, 2016) provided datasets for ABSA with a goal to foster research in this field. In this research, we have used the restaurant and laptop review datasets of SemEval'14\footnote{http://alt.qcri.org/semeval2014/task4}\cite{pontiki2014semeval}.
Restaurant domain dataset contains 3041 review sentences while Laptop domain dataset contains 3845 review sentences. All reviews are written in the English language. They also provided testing data (800 review sentences) separately for each domain. A snapshot of the dataset is given in Figure~\ref{FIG:SemEval14Dataset}. The statistics of the datasets is presented in Table~\ref{tab:dataset_summary}. In the dataset, review sentences have the labeled information of aspect-terms and aspect-term polarities for restaurant and laptop reviews. 

\begin{table}[!ht]
  \centering
  \caption{Dataset distribution}
  \label{tab:dataset_summary}
    {
    \begin{tabular}{lrrr}
    \toprule
    \textbf{Domain} & \textbf{Train} & \textbf{Test} & \textbf{Total} \\ \midrule
    Restaurants & 3041  & 800   & 3841 \\
    Laptops & 3045  & 800   & 3845 \\ \cmidrule(l){2-4}
     & 6086  & 1600  & 7686 \\ \bottomrule
    \end{tabular}}
\end{table}

For each entity, aspect-terms can pose any number of words.
Further, each review sentence can contain many aspect-terms.
We have filtered the reviews that do not contain any aspect-terms. 
Aspect-term polarity can be assigned to one of the four classes. 
These four classes are positive, negative, conflict (both positive and negative sentiment), and neutral (neither positive nor negative sentiment).
\begin{figure}[!ht]
 \centering
 \includegraphics[width=0.45\linewidth]{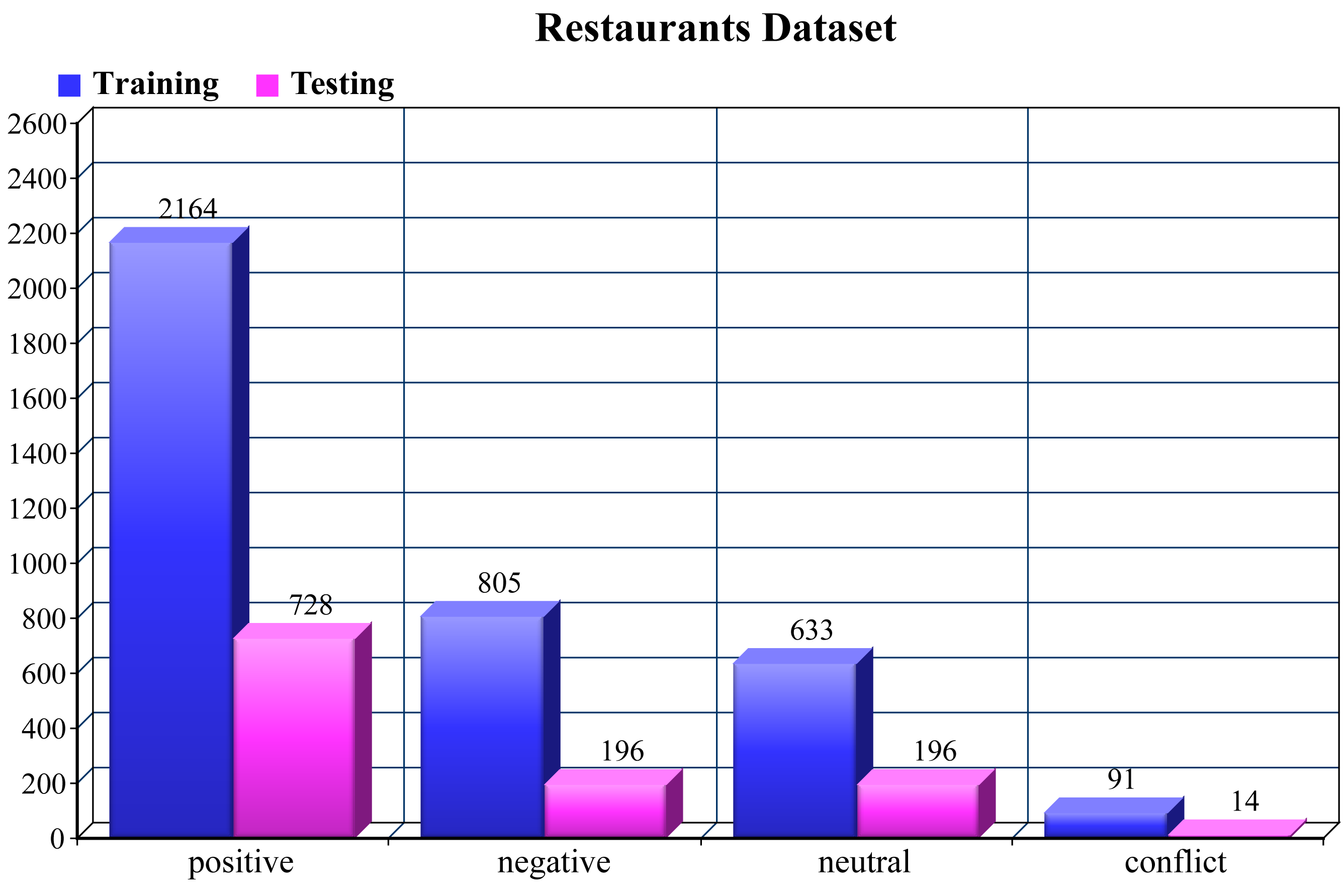}
 \includegraphics[width=0.45\linewidth]{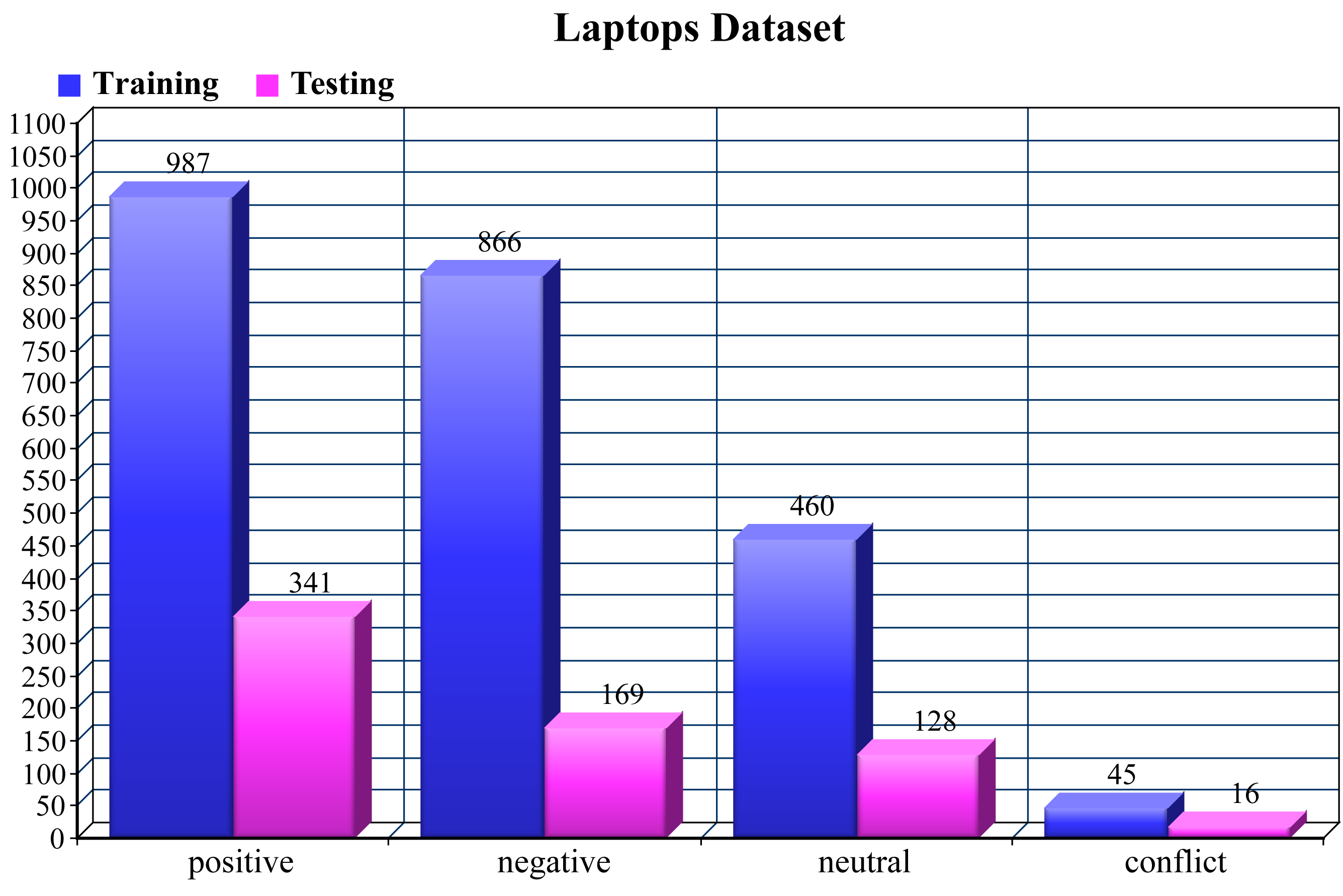}
 \caption{Aspect-terms sentiment polarity class distribution for Restaurants and Laptops dataset}
 \label{FIG:ATP_pol_dist}
\end{figure}
\subsection{Parameter Settings and Model Training}
To represent the input text as a dense vector, we have used ELMo embeddings for our experiments. We evaluate our method and baseline methods on SemEval'14 restaurant and laptop review dataset for aspect-term sentiment analysis tasks. We defined our problem as a multi-class classification problem. So the sentiment class having the highest predicted probability is selected corresponding to the concern aspect-term. For evaluation purposes, we used the average as an evaluation metric.

For our ATP model architecture, we employ a Bi-LSTM to learn the sequential relationship between the words of the review sentence. Inputs to the Bi-LSTM layer are contextualized word-embeddings. Further, we implement an attention mechanism to focus on the word that contributes more to determining the true sentiment polarity of the sentence. This attention layer learns the input text representation in the form of contextual vectors based on the importance of each context word. This context vector will be treated as a feature for sentiment classification. At last, we used a softmax classifier to classify the review sentence into one of the given sentiment classes (positive, negative, neutral, and conflict). Since the number of training instances belonging to the conflict category is very small (below 100) we have upsampled this category instances.

We have trained our model with the help of pre-trained contextualized word-embeddings (ELMo). These word-embeddings are originally learned on 1 Billion Word Benchmark\footnote{http://www.statmt.org/lm-benchmark/} for different dimensions. 
Out of these different dimensions, we have performed our experiments with 1024 dimensions word-embeddings.
All the sort sentences are padded to match the maximum sentence length in the training data.
For position embedding, we have randomly initialized the vectors and weights are updated during the training process. 
We have used the LSTM with cell size 150, due to which Bi-LSTM produced the hidden states with 300 (2*150) dimensions. 
The network is trained with a learning rate of 0.001 using the back-propagation algorithm with RMSprop optimizer.
We have also used a learning rate scheduler that reduces the learning rate as the epoch number increases. Inputs are passed to the model in the batches of size 128. To avoid over-fitting, a dropout and recurrent dropout having a rate of 0.5 is used in LSTM. The output dimension is set to 4 for sentiment polarity detection. For sentence tokenization and dependency tree generation, we have used Spacy. For the implementation, we have used TensorFlow\footnote{https://www.tensorflow.org/} and Google Colab\footnote{https://colab.research.google.com/}.

\subsection{Performance Evaluation}
We examine our method and baseline methods on SemEval'14 restaurant and laptop review dataset for aspect-term sentiment analysis tasks. We defined our problem as a multi-class classification problem. Therefore, the sentiment class having the highest predicted probability is selected corresponding to the given aspect-term. For evaluation purposes, we used accuracy as an evaluation metric.

\begin{table}[!ht]
  \centering
  \caption{Experimental results$^\dag$ of different methods on SemEval'14 dataset.
    }
  \label{tab:exp}
    {
    \begin{tabular}{lll} \toprule
    \textbf{Model} & \multicolumn{2}{c}{\textbf{Accuracy}}\\
    \cmidrule(l){2-3}
    & {Restaurant dataset} & {Laptop dataset} \\
    \midrule
        TD-LSTM  & 75.63\% & 68.13\% \\
        AE-LSTM  & 76.6\% & 68.9\% \\
        ATAE-LSTM  & 77.2\% & 68.7\% \\
        MemNet  & 80.95\%\scriptsize(MemNet9) & 72.37\%\scriptsize(MemNet7) \\
        IAN  & 78.6\% & 72.1\% \\
        RAM  & 80.23\% & 74.49\% \\
        PBAN  & 81.16\% & 74.12\% \\
        Coattention-MemNet  & 79.7\% & 72.9\% \\
        GANN  & 80.09\% & 72.21\% \\
        ReMemNN  & 79.64\% & 71.58\% \\
        \textbf{ATP (our-model)}  & \textbf{81.39\%} & \textbf{73.39\%} \\
        
        \bottomrule
    \end{tabular}}
    \\
    \dag \footnotesize{All results are reported from original articles except for TD-LSTM whose results are copied from ReMemNN}
\end{table}

Table~\ref{tab:exp} demonstrates the performance of our ATP method with baseline methods. We observe from it that for the restaurant domain our model works best among the baseline methods and for laptop domain ATP manages state-of-the-art performance. It is also capable of determining the correct sentiments of those sentences which are containing multiple aspect-terms.

\subsection{Model Analysis (Case Study)}
Apart from quantitative analysis which we have done from Table~\ref{tab:exp}, it is always better to have qualitative analysis for natural language models. The importance of attention mechanism is reflected in the table itself but case wise examples will also help to grasp the more insights of attention. We have picked some examples from the test data to diversify the effectiveness of attention and visualize the learned attention weight for a given sentence and target aspect. In the following figures, the higher value of attention weights are represented by darker color and lower value of attention weights are visualized by light color. In short, values of weights are represented by color gradation.

\begin{figure}[!ht]
 \centering
 \includegraphics[width=0.7\linewidth,height=3cm]{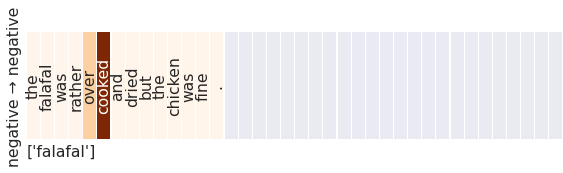}
 \includegraphics[width=0.7\linewidth,height=3cm]{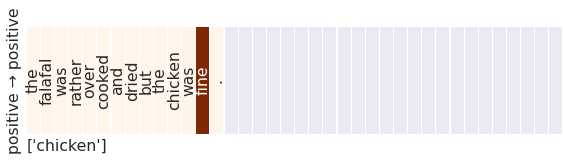}
 \caption{Example case 1}
 \label{FIG:ex1}
\end{figure}

\begin{figure}[!ht]
 \centering
 \includegraphics[width=0.7\linewidth,height=3cm]{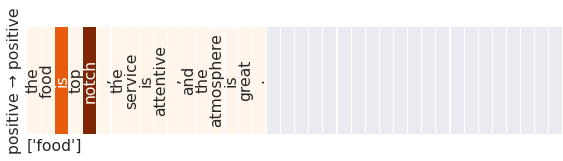}
 \includegraphics[width=0.7\linewidth,height=3cm]{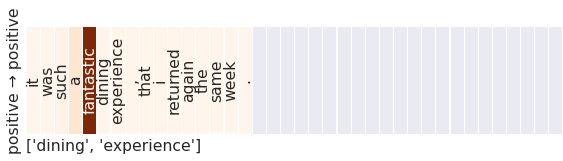}
 \caption{Example case 2}
 \label{FIG:ex2}
\end{figure}

In the first example, we have picked a sentence with two aspect targets \emph{falafal} and \emph{chicken} in Figure~\ref{FIG:ex1}. In Spite of having multiple target aspects presented in the sentence. Attention context changes according to the focus aspect and the model predicts the correct sentiment polarity when the focused aspect target changes.
In the second example, two sentences are illustrated in Figure~\ref{FIG:ex2}. In the first sentence, main sentiment determiner context is occurring after the target aspect \emph{food} while sentiment determiner context is occurring before the target aspect \emph{dining experience} in the second sentence. Attention weights in the figure are witness to model effectiveness. It attends the correct context irrespective of its position (left to aspect or right to aspect).

\begin{figure}[!ht]
 \centering
 \includegraphics[width=0.7\linewidth,height=3cm]{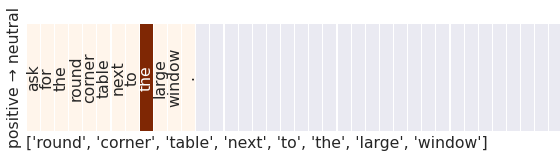}
 \caption{Example case 3}
 \label{FIG:ex3}
\end{figure}

Apart from the correct predictive examples, we have also chosen some examples where our model fails to determine actual sentiment. Most sentences that seem neutral are hard to predict by our model.
In Figure~\ref{FIG:ex3}, the actual polarity is positive but our model predicts it to neutral.

\begin{figure}[!ht]
 \centering
 \includegraphics[width=0.7\linewidth,height=3cm]{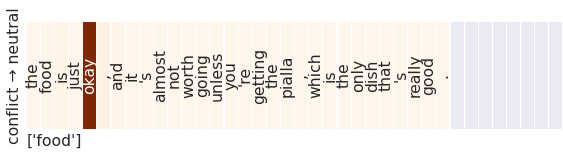}
 \includegraphics[width=0.7\linewidth,height=3cm]{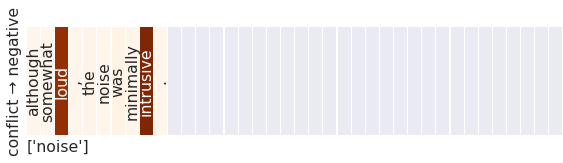}
 \caption{Example case 4}
 \label{FIG:ex4}
\end{figure}

Also, some conflict sentiment sentences are wrongly predicted by the model. In Figure~\ref{FIG:ex4}, sentences have conflict polarity but our model predict neutral and negative polarity respectively.

\section{Discussion and Conclusion\label{ATP_con}}
In this paper, we proposed an attention-based Bi-LSTM model ATP that is based on the position information. To incorporate the position information we use dependency trees. The key idea of using dependency trees to calculate the position information concerning a given aspect-term is to give importance to those significant words that are far apart based on the word-distance. If a word is connected to the aspect-term via dependency link then it has some relationship with the aspect-term and can play a significant role while determining the sentiment orientation. We performed experiments on SemEval'14 dataset and empirical results confirmed our assumption. 

As future work, we can consider trying to combining both word-distance and dependency path distance. We may also try other deep learning models to improve performance.

\bibliographystyle{unsrt}  
\bibliography{references}

\end{document}